%% file: main.tex
\documentclass[conference]{IEEEtran}
\IEEEoverridecommandlockouts

\usepackage{tikz}
\usepackage{cite}
\usepackage{amsmath,amssymb,amsfonts}
\usepackage{algorithm}
\usepackage{algpseudocode}
\algnewcommand\algorithmicforeach{\textbf{for each}}
\algdef{S}[FOR]{ForEach}[1]{\algorithmicforeach\ #1\ \algorithmicdo}
\usepackage{pifont}
\usepackage{graphicx}
\usepackage{subfigure}
\usepackage{textcomp}
\usepackage{multirow}
\usepackage{xcolor}
\usepackage{soul}

\usepackage[a4paper, total={184mm,239mm}]{geometry}
\usepackage{caption}
\def\BibTeX{{\rm B\kern-.05em{\sc i\kern-.025em b}\kern-.08em
    T\kern-.1667em\lower.7ex\hbox{E}\kern-.125emX}}

\begin{document}
\title{\vspace{-10pt}Enabling Fast Deep Learning on Tiny Energy-Harvesting IoT Devices\vspace{-10pt}}

\author{
    \IEEEauthorblockN{Sahidul Islam\IEEEauthorrefmark{1}, Jieren Deng\IEEEauthorrefmark{2}, Shanglin Zhou\IEEEauthorrefmark{2}, Chen Pan\IEEEauthorrefmark{3}, Caiwen Ding\IEEEauthorrefmark{2}, Mimi Xie\IEEEauthorrefmark{1}}
    \IEEEauthorblockA{\IEEEauthorrefmark{1}Department of Computer Science, The University of Texas at San Antonio
    }
    \IEEEauthorblockA{\IEEEauthorrefmark{2}Department of Computer Science, University of Connecticut
    }
    \IEEEauthorblockA{\IEEEauthorrefmark{3}Department of Computer Science, Texas A\&M University–Corpus Christi
    }
\vspace{-30pt}
}

\maketitle

\begin{abstract}

Energy harvesting (EH) IoT devices that operate intermittently without batteries, coupled with advances in deep neural networks (DNNs), have opened up new opportunities for enabling sustainable smart applications. Nevertheless, implementing those computation and memory-intensive intelligent algorithms on EH devices is extremely difficult due to the challenges of limited resources and intermittent power supply that causes frequent failures. To address those challenges, this paper proposes a methodology that enables fast deep learning with low-energy accelerators for tiny energy harvesting devices. We first propose $RAD$, a resource-aware structured DNN training framework, which employs block circulant matrix and structured pruning to achieve high compression for leveraging the advantage of various vector operation accelerators. A DNN implementation method, $ACE$, is then proposed that employs low-energy accelerators to profit maximum performance with small energy consumption. Finally, we further design $FLEX$, the system support for intermittent computation in energy harvesting situations. Experimental results from three different DNN models demonstrate that $RAD$, $ACE$, and $FLEX$ can enable fast and correct inference on energy harvesting devices with up to 4.26X runtime reduction, up to 7.7X energy reduction with higher accuracy over the state-of-the-art. 

\end{abstract} 

 
\maketitle


\input{Chapters/Introduction}

\input{Chapters/Background}
\input{Chapters/Algorithm}

\input{Chapters/Experiment}
\input{Chapters/Conclusion}
\bibliographystyle{abbrv}
\bibliography{main}

\end{document}

%% file: Chapters/Introduction.tex
\section{Introduction}\label{intro}
\vspace{-1pt}
In this IoT era where all things trend towards to be embedded with electronics, DNN brings new opportunities for embedded IoT devices to be smarter, more versatile, and able to handle more complex jobs. With DNN algorithms implemented on device, data can be processed locally instead of transmitting large amount of data to the cloud server for further processing.
Such a procedure not only significantly reduces processing pressure on the cloud but also improves energy efficiency and time for decision making.
Resource-constrained embedded IoT devices however face three major challenges from computation, memory, and energy perspectives for implementing DNNs. First, embedded IoT devices have significantly less computational units and lower CPU frequency (e.g. 1-16MHz) compared with high-performance computers. Since DNNs are computationally expensive, executing DNNs on embedded IoT devices results in significantly long execution time. Second, embedded IoT devices are equipped with much smaller (e.g. 1-16Kb) memory due to the size constraint. Third, majority IoT devices are powered with batteries, which can be depleted in days or even hours, requiring high maintenance cost which is especially undesirable if devices are implanted or deployed in harsh environment.

Energy harvesting (EH), which can scavenge energy from ambient environment, has become a promising technology to provide energy supply for IoT devices. However, the power supply can be intermittent and low. Since DNN execution is time and energy-consuming, the power outage is prone to happen frequently in the middle of the execution. As a result, implementing DNN in energy harvesting devices is a highly challenging job because the implementation requires consideration of both the resource constraints and the intermittent power supply which complicates the development of such systems. Recent work proposed software system SONIC and TAILS~\cite{intelBeyondEdge} to enable fast deep learning inference on battery-less systems. Those systems exploit the special structured and loop-heavy computations of DNN and enable their intermittent executions by continuously saving the loop control states to the non-volatile memory after each instruction. However, correctness of intermittent execution requires special mechanism to handle the inconsistency issues. 

To our knowledge, none of the existing work has considered the various vector operations supported by the low-energy accelerators in the training process and thus fail to efficiently leverage those accelerators in implementation and make efficient progress with existing software support for intermittent execution.
To generate the best framework for the DNN implementation, we need to design the DNN model based on the resource constraints and employ the compression methods that can take advantage of those hardware accelerators while considering their limitations and the constraints of device. Besides, to avoid the progress setback due to power failures while guaranteeing correctness, we need special support for those vector operations executed on accelerators.
To this end, this paper proposes the first framework that enables fast deep learning leveraging low-energy accelerators on resource-constrained EH IoT devices with three contributions as follows:


\begin{itemize}
  \item A resource-aware structured DNN training framework, $RAD$, is proposed that 
  consists of four different components: architecture search, compression, normalization, and fixed point calculation.
  \item A software system, $ACE$, is designed that efficiently implements DNN algorithm, 
  including acceleration-aware data flow design, hardware acceleration, overflow-aware computation and circular buffer convolution.
  \item A novel on-demand robust checkpointing scheme, $FLEX$, is proposed which can minimize or completely avoid progress setbacks, even if a power outage happens.
\end{itemize}
The remainder of this paper is organized as follows: Section~\ref{motiv} provides the background and related work; Section~\ref{technique} provides the framework overview and proposes the techniques for deep learning on energy harvesting devices; Section~\ref{exp} presents the experimental evaluation, and Section~\ref{conclude} concludes this work.

%% file: Chapters/Background.tex
\section{Background and Related Work}\label{motiv}

\noindent\textbf{Energy Harvesting Resource-Constrained Devices:}
Typical microcontrollers of IoT edge devices have minimal memory size and compute resources. 
Even a small DNN model like LeNet that has millions of operations takes several minutes to execute \cite{intelBeyondEdge}. As a result, implementing DNN algorithms is more challenging~\cite{zhou2021end}. DNN algorithms need to be redesigned and optimized to fit those small devices with good accuracy and run time. 
Besides, the energy needs to be harvested to a certain threshold to begin computation. However, the energy drained fast, resulting in frequent power failures, which incurs a problem with two dimensions. First, resuming the computation without wasted work. Second, checkpointing must not consume much energy and time. As a result, there are also opportunities to re-think about the traditional checkpointing mechanism.

\vspace{2pt}
\noindent\textbf{Low Energy Accelerators:}
To efficiently deploy DNN algorithms on energy harvesting devices, vector operation accelerators like TI's low-energy accelerator (LEA)~\cite{lea} can perform vector operations such as FFT(Fast Fourier Transform), IFFT (Inverse FFT), MAC (Multiply and Accumulation), ADD, etc. Such low energy vector processor performs vector operations without any CPU intervention.

\vspace{2pt}
\noindent\textbf{DNN Pruning:} Compared with other pruning solutions of DNN algorithms, block-circulant matrix (BCM) based DNNs~\cite{ding2017circnn} and structured pruning~\cite{yuan2019ultra}\cite{zhang2021unified} serves as good candidates for energy harvesting resource-constrained devices.

\noindent\subsubsection*{a) Structured Pruning} Structured pruning is proposed to structurally remove entire filters, channels, or filter shapes from the weight matrix~\cite{yuan2021tinyadc}. Structured pruning becomes more hardware-friendly by taking advantage of the pruned weight matrices with regular shapes~\cite{yuan2020dnn}, avoiding introducing extra indices that indicate the pruned locations. 

\noindent\subsubsection*{b) Block-circulant matrix}
Block-circulant matrix (BCM) based DNN algorithm can drastically reduce memory footprint and computation pressure, consequently achieving low energy consumption. Since its implementation requires Fourier and inverse Fourier transformation, we can employ the FFT/IFFT accelerators for implementation.

The key idea of block-circulant matrix (BCM) based fully connected (FC) layers is to partition the original weight matrix $\textbf{W}\in \mathbb{R}^{m\times n}$ into blocks of square sub-matrices, and each sub-matrix is a circulant matrix. Specifically, the matrix-vector multiplication can be implemented via ``FFT$\rightarrow$element-wise multiplication$\rightarrow$IFFT", i.e.,  $\text{IFFT}\big(\text{FFT}(\mathbf{p}_{ij})\circ\text{FFT}(\mathbf{x}_j)\big)$ using the BCM format.
For the inference phase, the computational complexity of this FC layer is $O(pqk\log k)$.
Similarly, the storage complexity is $O(pqk)$ because only $\mathbf{w}_{ij}$ or $\text{FFT}(\mathbf{w}_{ij})$ of each sub-matrix needs to be stored, which is equivalent to $O(n)$ for small $p$, $q$ values. 
Therefore, the simultaneous acceleration and model compression can be achieved. 

\begin{figure}[h]
\vspace{-4pt}\centering
\includegraphics[scale=0.25]{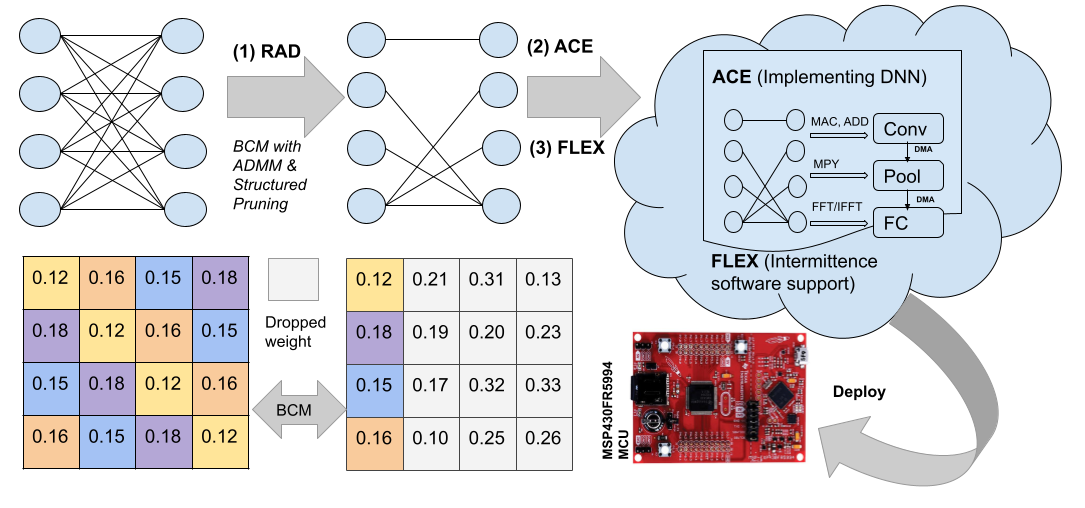}
\vspace{-15pt}
\caption{System Overview.}
\vspace{-16pt}
\label{Fig:System}
\end{figure}

\vspace{2pt}
\noindent\textbf{Related Works:} \label{realted} Several works have implemented CNN on IoT devices. SONIC is an intermittence-aware software system with specialized support for DNN inference~\cite{intelBeyondEdge}. NeuroZERO introduces a co-processor architecture consisting of the main microcontroller that executes scaled-down versions of a (DNN) inference task~\cite{neuroZero}. A software/hardware co-design technique that builds an energy-efficient low bit trainable system is proposed in~\cite{lowbit}. Efficient memory aware data flow designs for inference are proposed~\cite{memwarednn}.
Intermittent-Aware Neural Architecture Search for task based inference and Hardware Accelerated Intermittent inference is proposed in \cite{InNAS} and \cite{HAI}, respectively. ePerceptive is on-device CPU-based inference \cite{eperceptive}, based-on SONIC\cite{intelBeyondEdge}. 
Different from the above works, this is the first work that explores BCM-based DNN algorithms on resource-constrained energy harvesting IoT devices. 

%% file: Chapters/Algorithm.tex
\section{Deep Learning on Energy Harvesting Device}
\label{technique}
\noindent\textbf{System overview:} We propose a resource-aware training, pruning, and implementation framework that takes various types of vector operations supported by low-energy accelerators into consideration.  
Figure~\ref{Fig:System} shows the system overview of the proposed framework which consists of three techniques named $RAD$, $ACE$, and $FLEX$, as well as the responsibility of each technique, and their correlation. Resource-aware DNN training and pruning method, $RAD$, provides the resource-aware model, compression, normalization, and fixed point calculation. $ACE$ then implements DNN algorithm on EH device with efficient data flow design, circular buffer convolution, and hardware accelerators. A on-demand hybrid checkpointing method, $FLEX$, that also takes advantage of the special structure of DNN helps deal with the frequent power failures while avoiding progress setbacks.



\input{Chapters/Algorithm_modeling}

\input{Chapters/Algorithm_implementation}
\input{Chapters/Algorithm_energy}

\

%% file: Chapters/Algorithm_modeling.tex
\subsection{Resource-aware DNN Pruning ($RAD$)}
\begin{figure}[t]
\centering
\vspace{-0.1in}
\includegraphics[width=0.8\columnwidth]{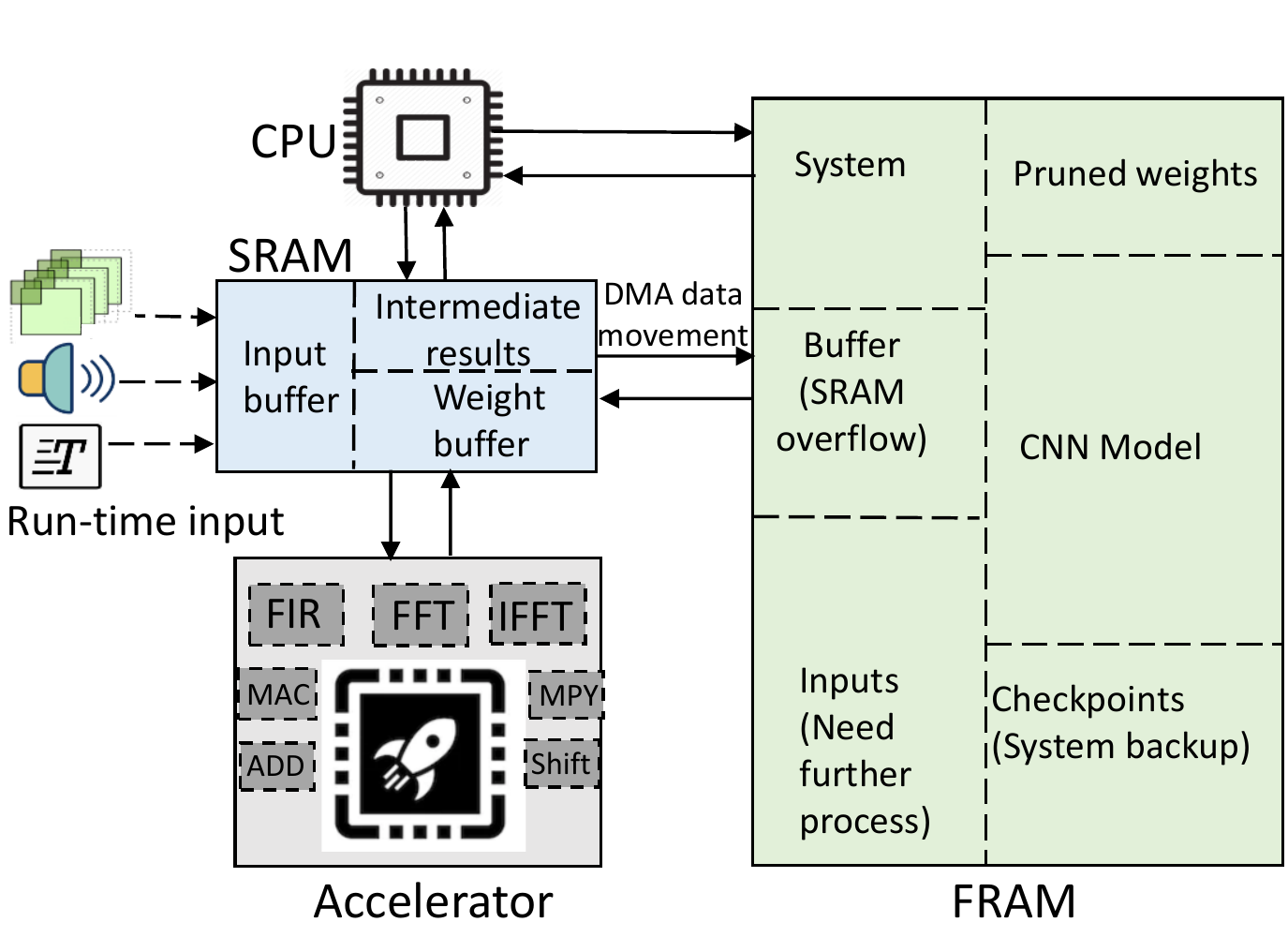}
 \vspace{-2pt}
\caption{Resource Overview in Embedded Devices.}
\vspace{-12pt}
\label{Fig:resource}
\end{figure}

$RAD$ aims to train and prune a DNN model that is aware of the available device resources. $RAD$ starts with a backbone model with good accuracy by doing architecture search. Then $RAD$ performs resource-aware compression with small accuracy loss while achieving less memory footprint and latency. $RAD$ trains the model offline.
Figure~\ref{Fig:resource} shows the typical computing and memory resources in an IoT device. Besides, CPU, LEA can be used for executing vector operations of DNN algorithms, while SRAM and FRAM are used as data storage. SRAM is smaller with lower dynamic read-write energy compared with FRAM. SRAM is used as a buffer for intermediate data, where FRAM is used for storing the DNN model and the control data for intermittent computation, which will be discussed in Section III.C.

\vspace{2pt}
\noindent\textbf{Modeling challenges:} \label{constraints}
There are several challenges during DNN modeling. 1) SRAM/FRAM size 2) CPU frequency and inference time 3) Overflow error 4) Model accuracy. 
Aware of the above challenges, $RAD$ trained the DNN model offline. The model must fit into the FRAM with acceptable inference time and accuracy. $RAD$'s architecture search technology finds a suitable model and further compresses it. As we deal with fixed-point calculation in a resource constraint device, there is possible data overflow during vector operations like MAC and FFT. As a result, $RAD$ performs several normalizations so that data stays between ranges during calculation.

\label{RAD}

\vspace{2pt}
\noindent\textbf{Fixed-point quantization:} Traditionally, DNN models are trained using high precision floating-point calculation. During on-device deployment, this is costly in terms of latency and memory. In this work, we use low-precision fixed-point representation.
To implement the low-precision fixed-point representation, $RAD$ maps high precision floating-point to 16-bits fixed-point so that $ACE$ can deploy the DNN model with fixed-point calculation.


\noindent\textbf{Normalization:} 
For normalization, $RAD$ first sets the data range $G_i$ with a minimum value, $G_{min}$, as -1 and a maximum value, $G_{max}$, as 1. $RAD$ then normalizes the data within this range of $[-1,1]$. Finally, to avoid the value of the computed intermediates exceeding this range during inference, $RAD$ uses the cosine normalization~\cite{luo2017cosine} to constrain the values of the computed intermediates into $[-1,1]$. 

\begin{table}[h]
\caption{BCM compression for 512*512 fully connected layer}
\label{BCM_memory_compression}
\begin{tabular}{|r|r|r|r|}
\hline
\multicolumn{1}{|l|}{\textbf{Kernel Size}} & \multicolumn{1}{l|}{\textbf{Block size}} & \multicolumn{1}{l|}{\textbf{Compressed kernel}} & \multicolumn{1}{l|}{\textbf{Storage reduction}} \\ \hline
 & 16 & 65536 Byte & 93.75\% \\ \cline{2-4} 
 & 32 & 32768 Byte & 96.87\% \\ \cline{2-4} 
 & 64 & 16384 Byte & 98.43\% \\ \cline{2-4} 
 & 128 & 8192 Byte & 99.21\% \\ \cline{2-4} 
\multirow{-5}{*}{\textit{1048576 Byte }} & 256 & 4096 Byte & 99.60\% \\ \hline
\end{tabular}
\vspace{-2mm}
\end{table}

\vspace{2pt}
\noindent{\textbf{Compression:} The fully connected (FC) layers are partitioned into 2D blocks of square sub-matrices and each of sub-matrices is a circulant matrix. This block circulant matrix (BCM)~\cite{ding2017circnn} enables the FFT-based computation for the FC layers. In addition, the structured pruning is also applied on the convoluitonal (CONV) layers to reduce the redundancy of CONV. The combination of BCM on FC and structured pruning on CONV can achieve drastically reduced memory footprint with a very small accuracy drop. Especially for the BCM, shown in Table~\ref{BCM_memory_compression}, the memory footprint reduction for FC layers is up to 99.60\%. }

\vspace{6pt}
\noindent \textbf{Alternating Direction Method of Multipliers (ADMM)-Regularized Optimization:}
The structured pruning problem can be effectively solved using the ADMM-regularized optimization. After the BCM-based pretrained DNN model is obtained, 
we define the constraint for the structured pruning separately on {\textit{accelerators}} via the ADMM-regularized  optimization~\cite{boyd2011distributed}. 

{\noindent Consider an $N$-layer DNN, where the weights and biases in the $i$-th CONV layer are ${\bf{W}}_{i}$ and ${\bf{b}}_{i}$. 
The optimization problem is shown below. 
}

\vspace{-1mm}
\begin{equation}
\small
\label{original}
\begin{aligned}
& \underset{ \{{\bf{W}}_{i}\},\{{\bf{b}}_{i} \}}{\text{minimize}}
& & \mathcal{F} \big( \{{\bf{W}}_{i}\}_{i=1}^N, \{{\bf{b}}_{i} \}_{i=1}^N),
\\ & \text{subject to}
& & 
{\bf{W}}_{i}\in {\bf{S}}_{i}, \; 
i=\{1,...,N\}
\end{aligned}
\end{equation}
\vspace{-0.5mm}
{\noindent where ${\bf{S}}_{i}$ = \{remaining non-zeros weights in convolutional layer satisfy the requirement of structured pruning\} is the pruning constraint set. 
We use the method proposed in ADMM-NN~\cite{ren2019admm} to solve the
problem.}

%% file: Chapters/Algorithm_implementation.tex
\subsection{Accelerator Enabled Embedded Software (ACE)}
\label{sec:task2-2}




To accelerate the on-device DNN inference for the pruned DNN model, we propose vector accelerator-enabled embedded software ($ACE$), which deploys DNN algorithms efficiently and ensures full utilization of the vector operations provided by the accelerators, such as FFT, IFFT, MAC, etc. 
Instead of element-wise computation, $ACE$ restructures the DNN computation by approaching bulk movement and vector operations in accelerators for both convolution and fully connected layer. Moreover, accelerator-based computation is susceptible to data overflow and executes only with fixed-point quantization. Therefore, $ACE$ performs overflow-aware computation with 16 bit quantization. In addition, $ACE$ proposes circular buffer convolution technique to reduce the memory footprint where the input/output memory is reused continuously.




\begin{figure*}[t]
\hspace{-0.25in}
\begin{center}
	\includegraphics[width = 0.8\textwidth, height=3.1cm]{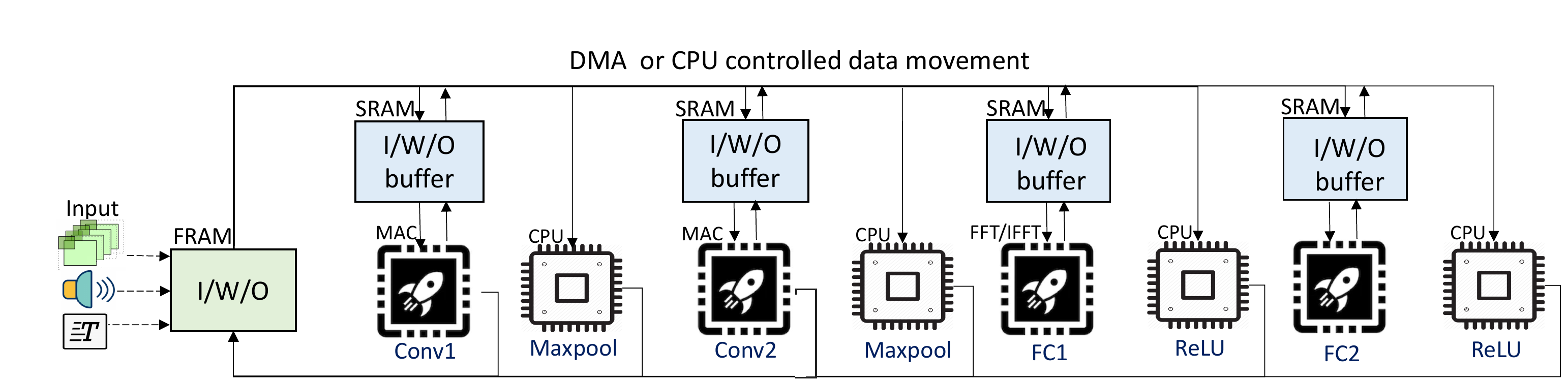}
      \vspace{-4pt}
	\caption{An example of on-device DNN data flow using LeNet-5.}
	\label{Fig:flow}
\end{center}
\vspace{-0.25in}
\end{figure*}

\vspace{2pt}
\noindent\textbf{Acceleration-aware dataflow:} 
Figure~\ref{Fig:flow} demonstrates the on-device data flow of LeNet-5 model as an example. Before executing each layer, the input and kernel are buffered in the SRAM allocated specifically for the accelerators. The buffered vectors are then sent to the accelerator for computation. Next, the outputs from the accelerators, buffered in SRAM, are stored in the FRAM. The subsequent Maxpooling layer and activation function (ReLU) is loaded directly into the CPU without SRAM buffering. 
$ACE$ also selects the right kind of data movement method based on the energy and latency of moving the data. For example, large vector of data is moved with DMA while a single data is moved with CPU. During the inference, more time is spent on data movement, thus utilizing DMA with bulk data transfer achieves significant improvement over CPU-based data transfer.

\noindent\textbf{Quantization:} $ACE$ provides software support for 16-bit fixed-point quantization to represent the floating-point data. $B => A*2^{b-1}$ is the common quantization rule for embedded devices where ``A" is a floating-point number, ``B" is a quantized number, and ``b" represents quantization bit.

\begin{algorithm}[h]
\tiny
\caption{On-device BCM implementation}
\footnotesize
\begin{algorithmic}[1]

    \State $\textbf{Input: } W,I$
    \State $\textbf{Output: } O$
    \State $I \gets $SCALE-DOWN$(I,len(I))$ 
    \State $W \gets $SCALE-DOWN$(W,len(W))$
    \State $cI \gets  $COMPLEX$(I)$ 
    \State $cW \gets  $COMPLEX$(W)$
    \State $cOut \gets  $IFFT$($FFT$(cI)*$FFT$(cW))$ 
    \State $O \gets  $REAL$(cOut)$ 
    \State $O \gets  $SCALE-UP$(O,len(I),len(W))$
    \State \Return O
\Procedure{Scale-down}{$\mathcal D$, $length$}
    \ForEach{$element \in \mathcal D $}
        \State $element \gets element/length$  
    \EndFor
    \State \Return $\mathcal D$
\EndProcedure
\Procedure{Scale-up} {$\mathcal D$, $lI$, $lW$}
    \ForEach {$element \in \mathcal D $}
            \State $element \gets element*lI*lW$  
    \EndFor
    \State \Return $\mathcal D$
\EndProcedure
\end{algorithmic}
\end{algorithm}

\vspace{2pt}
\noindent\textbf{Hardware Acceleration of FC layer:} 
With the FFT/IFFT low energy vector operations, the DNN inference time and energy can be reduced significantly for BCM-based fully connected (FC) layers.
We redesign the computation of FC layer as shown in  Algorithm 1, which describes the on-device implementation of BCM computation. In lines 3 and 4, input and weight data are scaled-down to overcome FFT operation overflow. Since data is not in complex form during the convolution, the real input needs to be converted into the complex number before performing FFT operation. $ACE$ performs complex/real operations in lines 5-8. Line 7 performs the core function of the BCM based DNN. It performs FFT, complex number multiplication on both inputs, and then IFFT on the result. After that, real number is retrieved at line number 8 and stored to FRAM. In Line 9, output data is scaled up.

\begin{figure}[h] 
\centering
\includegraphics[scale=0.55]{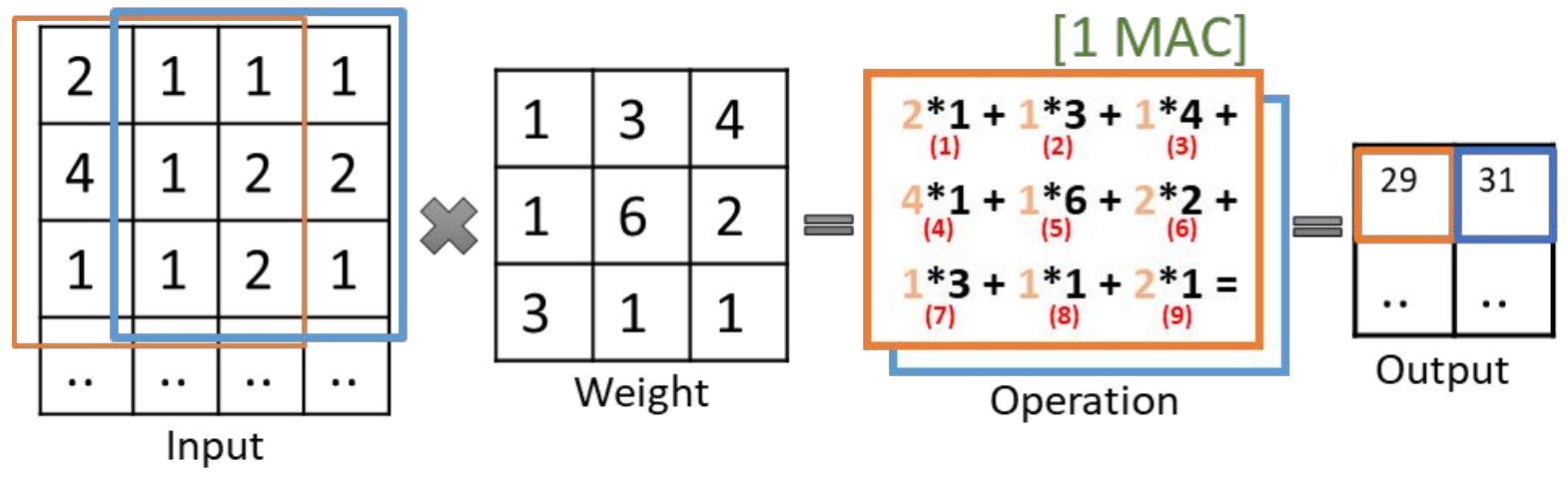}
\vspace{-0.22cm}
\caption{Example of MAC Operation}
\vspace{-0.4cm} 
\label{Fig:conv}
\end{figure}

\vspace{2pt}
\noindent\textbf{Hardware Acceleration of CONV layer:}
Instead of element-wise computation, $ACE$ implements \textit{bulk computation} by computing the whole kernel at a time. For example, as Figure \ref{Fig:conv}, a typical convolution of a 3x3 kernel needs  18 operations (9 multiplication and 9 addition) with an input window. Instead, an accelerator can replace the whole with a single MAC \cite{lea} operation, resulting in drastic improvement in overall convolution performance.

\noindent\textbf{Overflow-aware Computation:} Fixed point calculation in resource constraint devices frequently suffers from data overflow errors. During FFT computation, there is a possible data overflow if the totaling of the input array exceeds the capacity of the quantized bit. 
For example, in an 8-bit data quantization, the FFT will produce wrong results if the addition of the input array elements exceeds $2^8=256$. To address the challenge, $ACE$ performs data scaling based on the array size in Algorithm 1. 
Besides, there is also possible overflow during accelerator operation. The solution is provided using normalization as discussed in Section \ref{constraints}.

\begin{figure}[h] 
\centering
\vspace{-10pt}
\includegraphics[width=0.6\columnwidth, height=4cm]{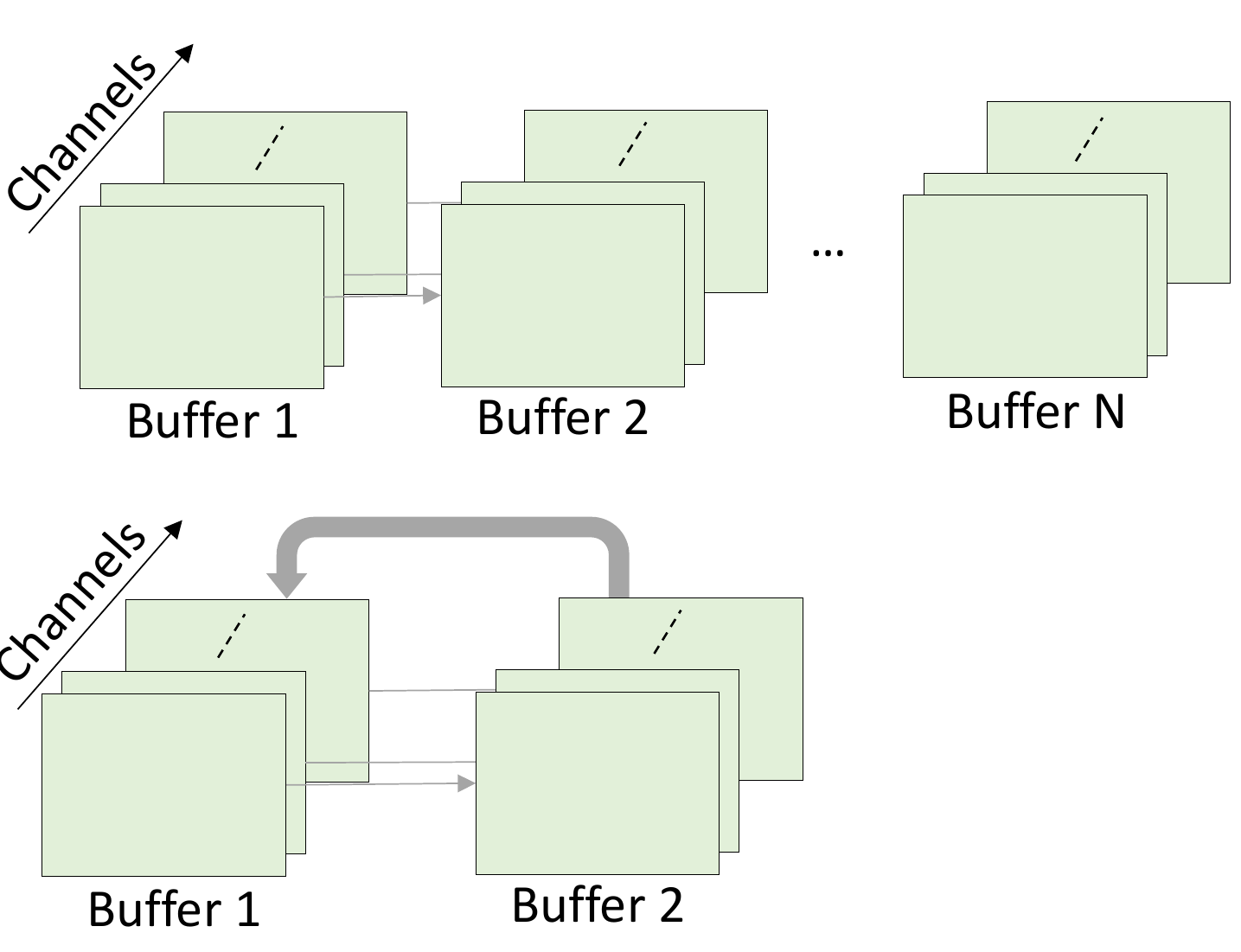}
 \vspace{-6pt}
\caption{Circular buffer for input/output buffer reuse.}
 \vspace{-10pt}
\label{Fig:buf}
\end{figure}

\vspace{2pt}
\noindent\textbf{Circular Buffer Convolution:} $ACE$ optimizes the memory usage by reusing the input/output buffer after a layer-level computation takes place. In the existing DNN implementation, an inference with $N$ layers requires N buffers, as shown in figure~\ref{Fig:buf}. Instead of allocating memory for individual layers, $ACE$ requires only two buffers (input and output) at most. $ACE$ implemented circular buffer convolution, restructured the DNN inference, which reuses the buffer by interchanging and overwriting the input and output pointer after finishing a layer-level computation. $ACE$ can considerably reduce the memory footprint, irrespective of the layer size in a DNN inference. The size required for the buffer is $max(L_{i}),1 \leq i \leq N $.



%% file: Chapters/Algorithm_energy.tex
\subsection{Intermittent Inference with $FLEX$.}
\label{sec:task1-3}
We designed $FLEX$, a software support that enables intermittent inference when there are frequent power failures.
Previous work can successfully checkpoint the system state, which, however, does not work efficiently due to the large amount of intermediate results and computation state that takes much time and energy. 
A recent work~\cite{intelBeyondEdge} proposes $TAILS$ that exploits the special structured and loop-heavy computations of DNN and enables their intermittent executions by continuously saving the loop control states. 
However, it will cause progress setbacks when FFT based computation is introduced.


\begin{figure}[h]
 \vspace{-0.18in}
\begin{center}
	\includegraphics[width = 0.5\textwidth]{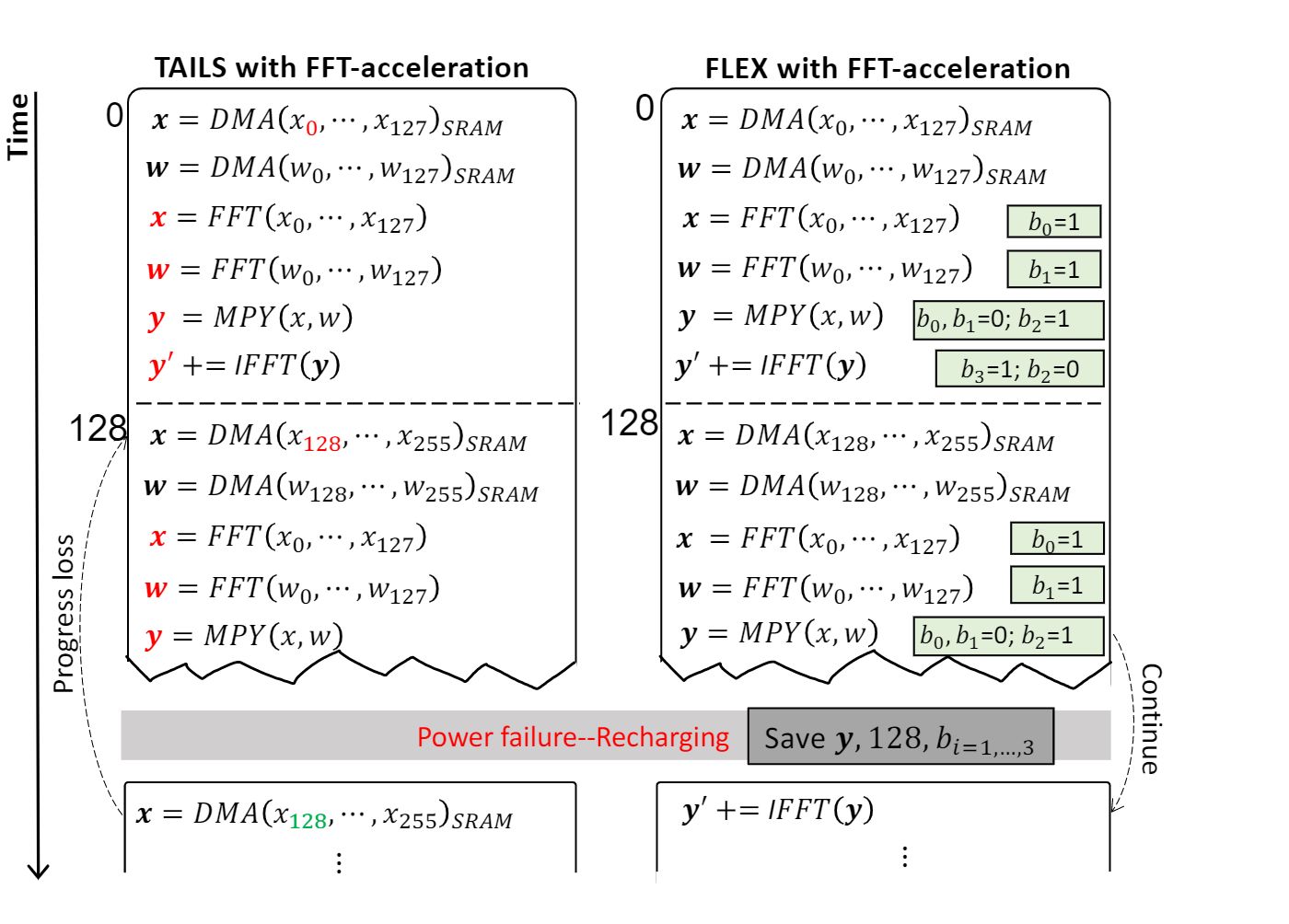}
	\vspace{-10pt}
	\caption{Comparison between TAILS and FLEX.}
	\label{Fig:flex}
\end{center}
\vspace{-18pt}
\end{figure}


\vspace{2pt}
\noindent\textbf{BCM-based FC layer:} The left part of Figure~\ref{Fig:flex} demonstrates the drawback of $TAILS$ when we try to employ FFT-based BCM computation under the intermittent power supply. An FFT-based BCM computation has to go through several steps including DMA, FFT/IFFT, and MPY as shown in algorithm 1. The arrays highlighted in red, \textbf{$x, w, y, y'$} are intermediate values. In loop-index based techniques, these are very much susceptible to data loss whenever a power failure happens. On power restore, the program has to roll back to the initial $DMA$ operation because only the loop index is insufficient to retrieve the current state of the Algorithm. It causes wasted work and hurts the progress. As a solution, $FLEX$ stores block index, intermediate result and state bit of the Algorithm. As shown in the right part of Figure~\ref{Fig:flex}, $FLEX$ specifies the 4 additional bits $b_0$-$b_2$ to indicate the current state, resulting in a successful continuation from the interrupted operation on power failure. As the control bits are small, it requires small memory footprint and read/write energy cost.

\noindent\textbf{Other layer:} 
In BCM-based FC layers, with the help of a voltage monitor system, $FLEX$ predicts a power failure and checkpoint the latest intermediate result. Apart from that, for all other layers, $FLEX$ follows loop index-based checkpointing mechanism that checkpoint and restores the system, based on loop indices.

\vspace{-12pt}

%% file: Chapters/Experiment.tex
\subsection{Experimental Setup}
\noindent\textbf{Hardware Setup:}
The proposed DNN implementation framework is evaluated with TI's MSP430FR5994 \cite{msp430fram} ultra-low-power evaluation board, consisting of a 16 MHz MCU, a 8KB volatile SRAM, a 256KB nonvolatile FRAM memory, and a low-energy Accelerator (LEA) that runs independently of CPU. 
The LEA accelerator supports FFT, IFFT, MAC, and MPY operation. 
The board is connected to the function generator SIGLENT SDG1032X~\cite{sdg1032x} to simulate the energy harvesting scenario. Energy is buffered with a capacitor of 100µF. For the energy measurement, we used CCS energy trace technology~\cite{EnergyTrace}.

\begin{table}[t]
\caption{Structure and Accuracy of DNN}\label{tab:models}
\centering
\resizebox{1\columnwidth}{!}{%
\begin{tabular}{|clcccc|}
\hline
\textbf{Tasks}                                                                              & \multicolumn{1}{c}{\textbf{Layer}} & \textbf{Original Size} & \textbf{Compress Method} & \textbf{Compression} & \textbf{Accuracy}      \\ \hline
\multirow{4}{*}{\begin{tabular}[c]{@{}c@{}}MNIST\end{tabular}}     & Conv                               & 6 x 1 x 5 x 5              & —                           & —                          & \multirow{4}{*}{99\%}  \\
                                                                                            & Conv                               & 16 x 6 x 5 x 5             & Structured Pruning              & 2x                       &                        \\
                                                                                            & FC                                 & 256 x 256                  & BCM                         & 128x                       &                        \\
                                                                                            & FC                                 & 256 x 10                   & —                           & —                          &                        \\ \hline
\multirow{4}{*}{\begin{tabular}[c]{@{}c@{}}HAR\end{tabular}} & Conv                               & 32 x 1 x 1 x12             & —             & —                       & \multirow{4}{*}{89\%} \\
                                                                                            & FC                                 & 3520 x 128                 & BCM                          & 128x                          &                        \\
                                                                                            & FC                                 & 128 x 64                   & BCM                         & 64x                       &                        \\
                                                                                            & FC                                 & 64 x 6                     & —                       & —                       &                        \\ \hline
\multirow{6}{*}{\begin{tabular}[c]{@{}c@{}}OKG\end{tabular}}         & Conv                               & 6 x 1 x 5 x 5              & —              & —                         & \multirow{6}{*}{82\%} \\
                                                                                            & FC                               & 3456 x 512              & BCM              & 256x                       &                        \\
                                                                                            & FC                                 & 512 x 256                  & BCM                         & 128x                      &                        \\
                                                                                            & FC                                 & 256 x 128                  & BCM                         & 64x                       &                        \\
                                                                                            & FC                                 & 128 x 12                  & —                       & —                       &                                              \\ \hline
\end{tabular}}
\label{models}
\end{table}

\begin{figure*}
\vspace{-13pt}
\hfill
\subfigure[Inference time on continuous power]{\includegraphics[width=7.1cm,height=3.4cm]{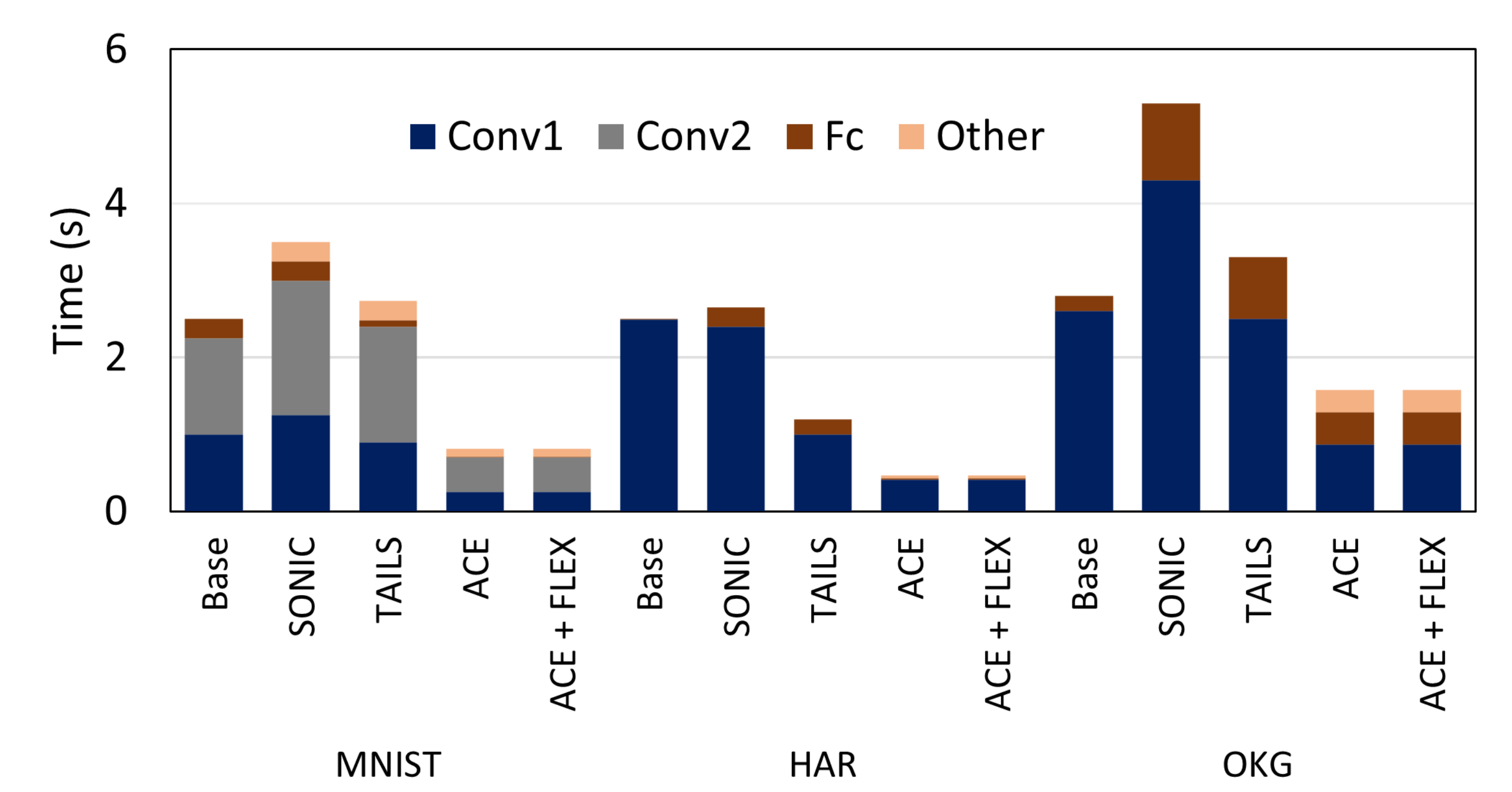}\label{different_dnn_runtime_on_power}}
\hfill
\hspace{-12pt}
\subfigure[Inference time on intermittent power (100µF)]{\includegraphics[width=5.8cm,height=3.4cm]{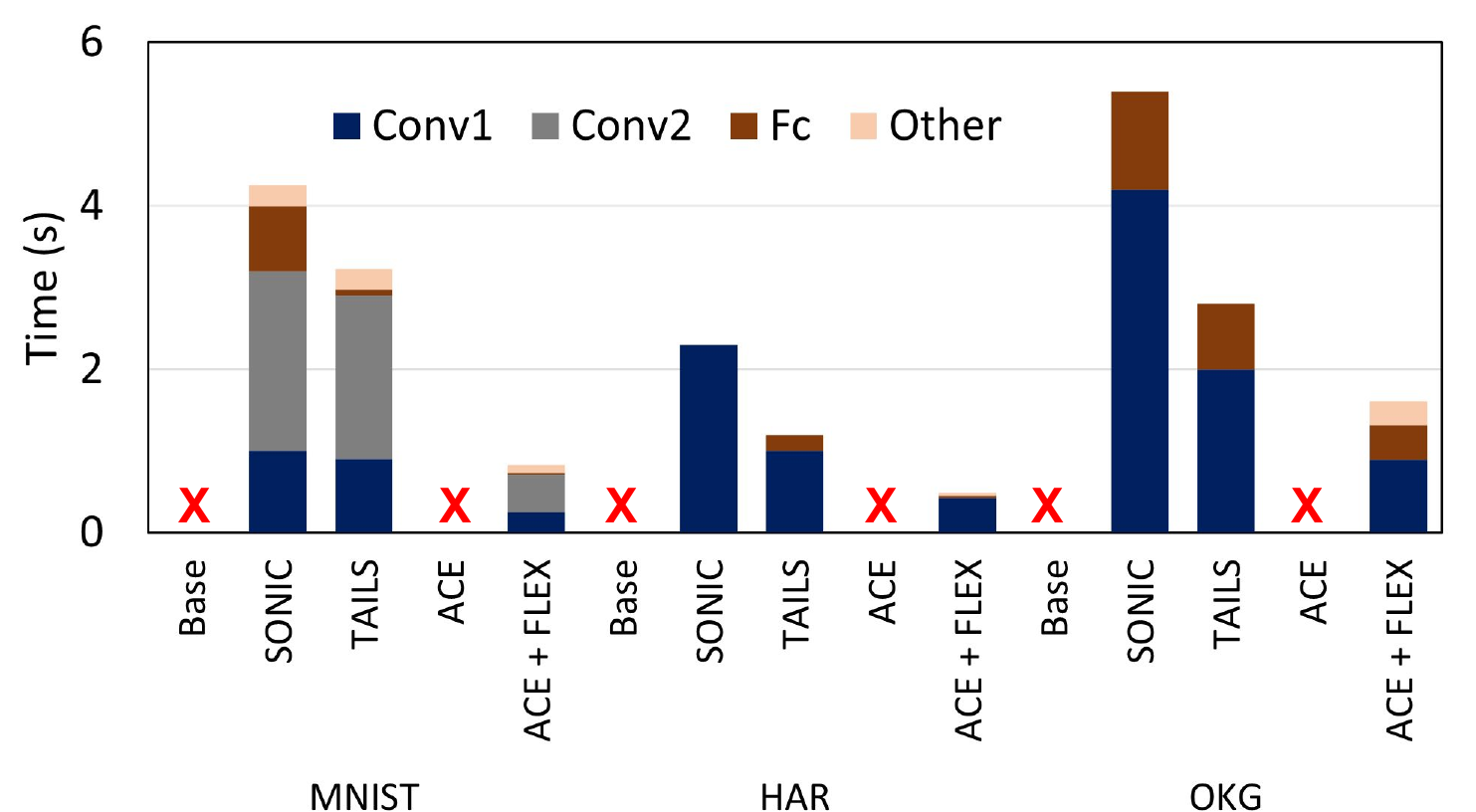}\label{dnn_runtime_on_intermittent_power}}
\hfill
\hspace{-12pt}
\subfigure[Energy breakdown]{\includegraphics[width=5.4cm,height=3.40cm]{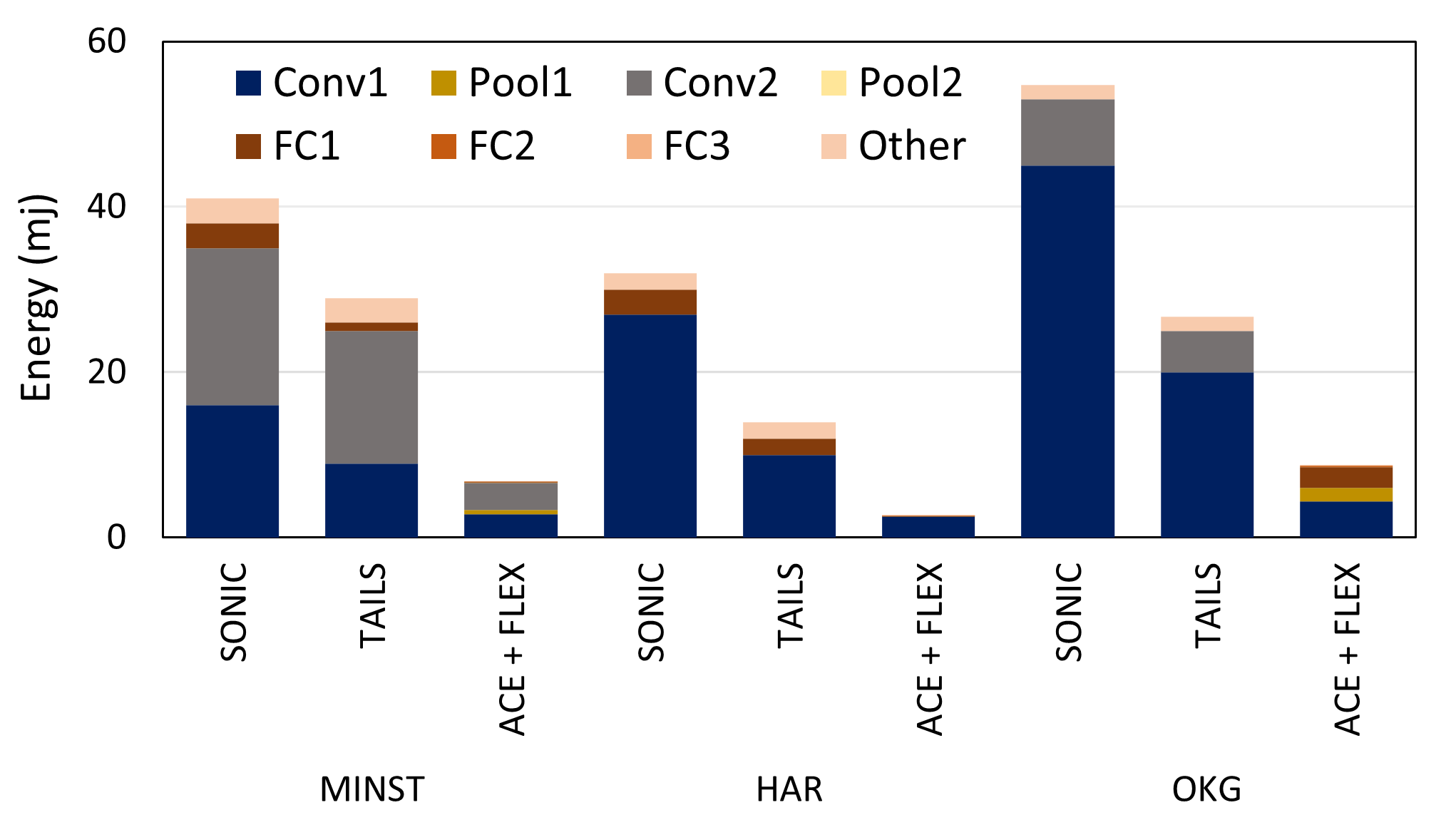}\label{different_dnn_energy}}
\hfill
\vspace{-6pt}
\caption{Comparison of DNN Implementation with Different Framework}
\vspace{-5pt}
\label{dif_dnn}
\end{figure*}

\begin{figure}
\vspace{-15pt}
\hfill
\subfigure[Latency]{\includegraphics[width=4.4cm, height=2.6cm]{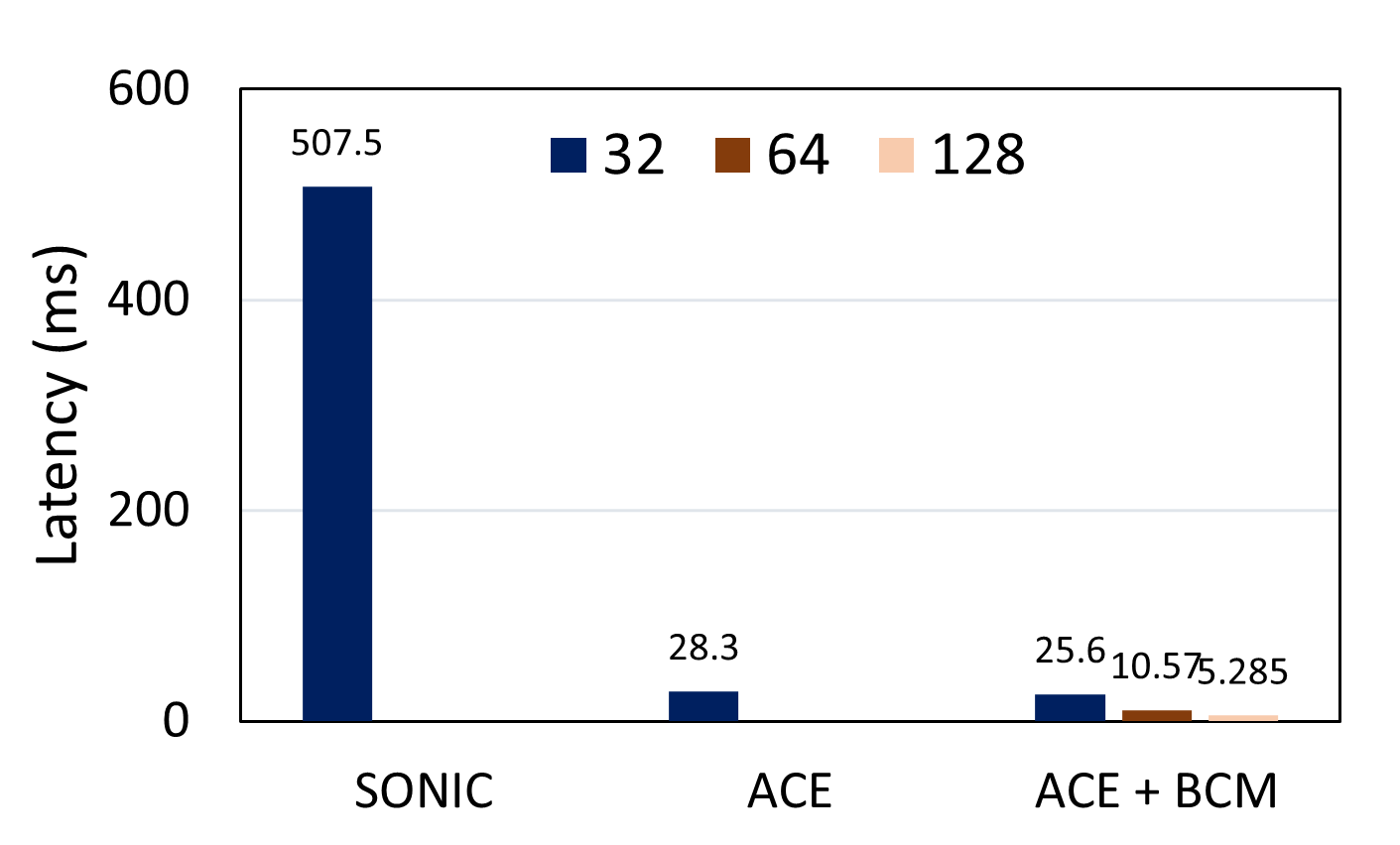}\label{FC_performance}}
\hfill
\hspace{-15pt}
\subfigure[Energy]{\includegraphics[width=4.4cm, height=2.6cm]{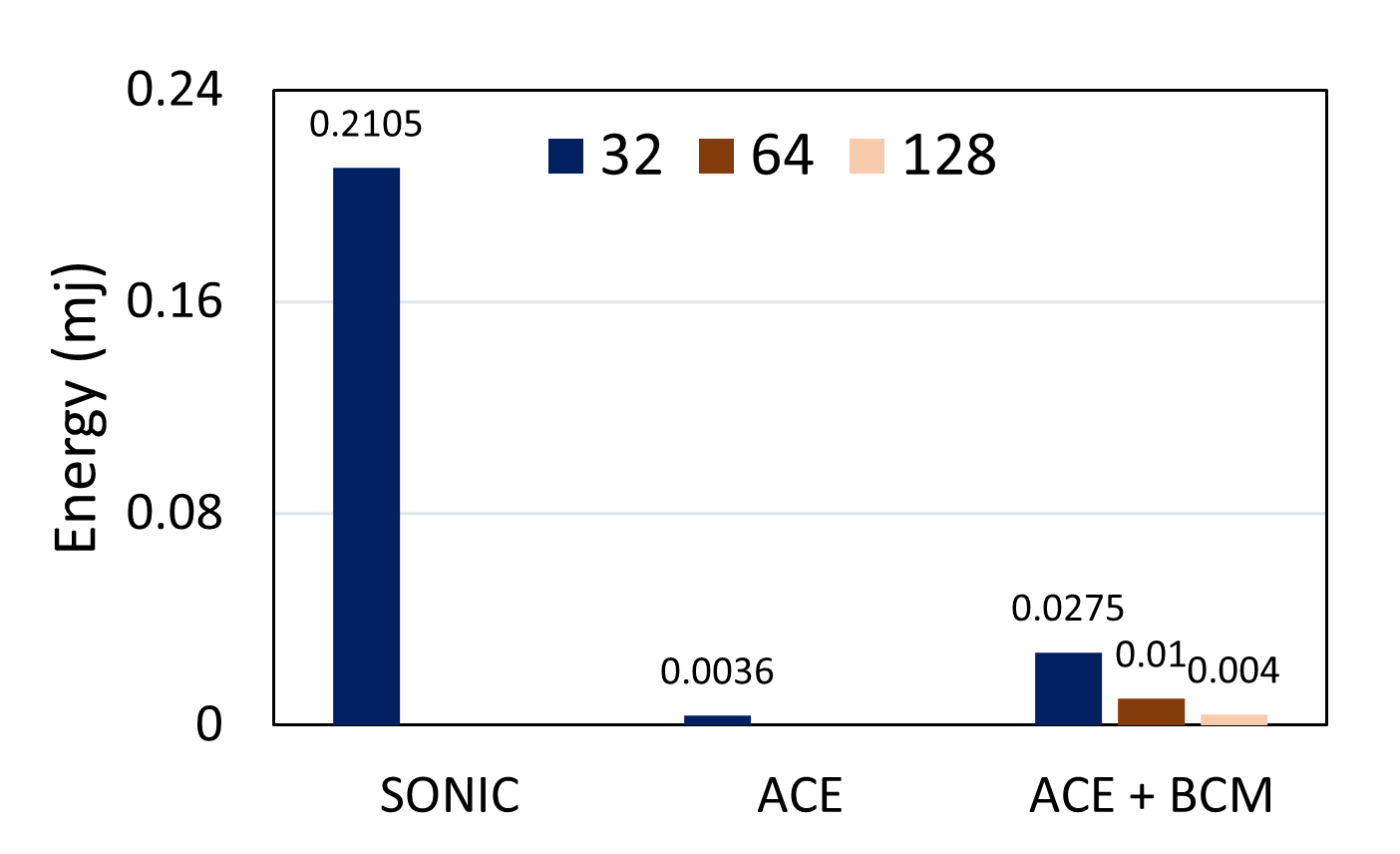}\label{FC_energy}}
\hfill
\vspace{-6pt}
\caption{Performance of First FC of MINIST.}
\label{FC_performance_en}
\vspace{-18pt}
\end{figure}

\section{Experiments}\label{exp}


\vspace{3pt}
\noindent\textbf{DNN Models:}
This paper considers three DNN models for Image Classification (MNIST)~\cite{lecun-mnisthandwrittendigit-2010}, Human Activity Recognition (HAR)~\cite{anguita2013public_HAR}, and Google Keyword Recognition (OKG)~\cite{google_keyword} which represent image-based applications, wearable applications, and audio applications respectively as shown in Table~\ref{tab:models}.

\subsection{Experimental Results}
\vspace{-3pt}
We evaluated our framework by comparing with BASE, SONIC and TAILS~\cite{intelBeyondEdge}, the state-of-the-art solutions. Here, BASE is the baseline implementation that does not consider intermittent operation.

\subsubsection{DNN Model Training and Pruning} 
As shown in Table~\ref{tab:models}, the structured pruning reduces the number of weights for the CONV layer by 2x and the BCM reduces the weights of the FC layer by 128x for the DNN model on MNIST task. For the HAR task, BCM reduces the number of weights by 128x for the first FC layer and the number of weights by 64x for the second FC layer. More FC layers of DNN models on the OKG dataset are compressed via BCM. The weights of the first three FC layers on this model are reduced for for 256x, 128x and 64x respectively.

\subsubsection{Inference Time under Continuous Power Supply}
We can observe from the figure~\ref{different_dnn_runtime_on_power} that $ACE$ and $FLEX$ run 3x, 4x, 3.3x faster inference than the Base, SONIC, and TAILS, respectively, on MNIST dataset. For HAR dataset, it is 5.4x, 5.7x, 2.6x faster. And lastly, it performs 1.7x, 3.3x, 2.1x faster on OKG dataset. This significant performance improvement is achieved by allowing the on-board acceleration engine to perform at the full extent with DMA-based data transfer. During the inference, we can see that most of the computation time is spent on convolutional layer while FC layer runs extremely fast.

\subsubsection{Inference Time under Intermittent Power Supply} Under intermittent power supply, as shown in Figure ~\ref{dnn_runtime_on_intermittent_power}, the Base model and $ACE$ can never be completed because they do not tolerate intermittence. However, $ACE$ with $FLEX$ can successfully complete the inference because $FLEX$ is developed to provide intermittent support. Besides, $FLEX$ also reduced the wasted work and checkpointing overhead. Due to power failure, there is a negligible increase (1\%-2\%) in latency and energy consumption, achieving almost similar latency and energy as continuous power.
Here, $ACE$ + $FLEX$ run 5.1x, 3.8x faster than the SONIC, and TAILS, respectively on MNIST dataset. For HAR dataset, it is 4.7x, 2.4x faster. And lastly, it performs 3.3x, 1.7x faster on OKG.

\subsubsection{Energy Consumption and Performance}
In terms of energy consumption, our techniques performed ever better. We can observe from figure~\ref{different_dnn_energy} that $ACE$ and $FLEX$ outperform SONIC and TAILS by achieving 6.1x and 4.31x energy-saving on MNIST. 10.9x and 5.26x energy-saving on HAR. And finally, 6.25x and 3.05x energy is saved on OKG. LEA and DMA run in ultra-low power mode and resulting in a significant amount of energy saving.

Specifically, Figure \ref{FC_performance_en} demonstrates how $ACE$ is successful in improving the overall performance of an FC layer. $ACE$ is performed both independently and with different BCM block sizes (32, 64, 128) on the first FC layer of MNIST model. The employment of BCM based DNN inference proposed in this paper can significantly reduce the latency and energy needed. A larger block size of BCM reveals better performance and more compression. However, selecting a larger block size is limited by device support and accuracy degradation, which proves the efficiency of our resource-constrained framework.

\subsubsection{Evaluation of checkpointing overhead}
The proposed checkpointing mechanism considers special structure of DNN models and thus only saves latest intermediate result, necessary loop indices, and 4 control bits as shown in Figure~\ref{Fig:flex}. Every checkpoint/restore cost is at most 0.033mj, which is reached if power failure happens when computing the FFT-based BCM in the FC layer. Therefore, in our experiment, the total checkpoint/restore overhead is only 1\%, 1.25\%, and 0.8\% for MNIST, HAR, and OKG datasets, respectively, which is negligible considering the largely overall saved energy.

%% file: Chapters/Conclusion.tex
\section{Conclusion}\label{conclude}
This paper proposes an efficient framework for DNN implementation on energy harvesting devices which includes a resource-aware DNN training and pruning method considering hardware resource constraints, a DNN implementation method utilizing accelerators, and software support for intermittent computation that's friendly to long vector operations with accelerators. Specifically, the BCM compression is successfully implemented with the FFT vector operation accelerators and the proposed quantization method which can improve the performance of FC layers by tens of times. The experimental results demonstrate significantly reduced DNN footprint, energy consumption, and improved performance.